# DC-AL GAN: Pseudoprogression and True Tumor Progression of Glioblastoma Multiform Image Classification Based on DCGAN and AlexNet

Meiyu Li, Hailiang Tang, Michael D. Chan, Xiaobo Zhou, and Xiaohua Qian

*Abstract*—Pseudoprogression (PsP) occurs in 20-30% of patients with glioblastoma multiforme (GBM) after receiving the standard treatment. In the course of post-treatment magnetic resonance imaging (MRI), PsP exhibits similarities in shape and intensity to the true tumor progression (TTP) of GBM. So, these similarities pose challenges on the differentiation of these types of progression and hence the selection of the appropriate clinical treatment strategy. In this paper, we introduce DC-AL GAN, a novel feature learning method based on deep convolutional generative adversarial network (DCGAN) and AlexNet, to discriminate between PsP and TTP in MRI images. Due to the adversarial relationship between the generator and the discriminator of DCGAN, high-level discriminative features of PsP and TTP can be derived for the discriminator with AlexNet. Also, a feature fusion scheme is used to combine higher-layer features with lower-layer information, leading to more powerful features that are used for effectively discriminating between PsP and TTP. The experimental results show that DC-AL GAN achieves desirable PsP and TTP classification performance that is superior to other state-of-the-art methods.

## I. Introduction

GLIOBLASTOMA multiforme (GBM) is one of the most common brain tumors, and is primarily caused by the canceration of glial cells in the brain and the spinal cord.

M. Li is with College of Electronic Science and Engineering, Jilin University, Changchun 130012, China, and also with Institute for Medical Imaging Technology, School of Biomedical Engineering, Shanghai Jiao Tong University, Shanghai 200030, China.

H. Tang is with Department of Neurosurgery, Huashan Hospital, Fudan University, Shanghai 200040, China.

M. Chan is with Department of Radiology, Wake Forest School of Medicine, Winston-Salem, NC 27157, USA.

X. Zhou is with School of Biomedical Informatics, The University of Texas Health Science Center at Houston, Houston, TX 77030, USA.

X. Qian is with Department of Radiology, Wake Forest School of Medicine, Winston-Salem, NC 27157, USA, and also with Institute for Medical Imaging Technology, School of Biomedical Engineering, Shanghai Jiao Tong University, Shanghai 200030, China (correspondence e-mail: Xiaohua.qian@sjtu.edu.cn).

Currently, the standard treatment of GBM includes surgery, radiotherapy, chemotherapy, and targeted therapy [1]. Among GBM patients who have received routine treatment, the probability of occurrence of pseudoprogression (PsP) is about 20 to 30% [2]. PsP is a subacute symptom that mimics truetumor progression (TTP) at the tumor site or resection margins, but PsP subsequently regresses or remains stable [3, 4]. The differentiation between PsP and TTP in clinical practice is mainly based on analyzing visible changes in the MRI of the lesion area. However, such analysis is typically time-consuming, can cause missing a patient's optimal treatment time, and has hence the treatment delay can have a detrimental effect on the treatment outcomes. In addition, biopsy of brain tumors isn't widely recommended due to the invasiveness and increased risks of the procedure. In summary, it is necessary to find a better noninvasive and efficient method to distinguish between PsP and TTP of GBM.

Over the past decade, researchers have devoted considerable efforts to explore methods based on genetic and molecular markers, as well as image features for differentiating PsP and TTP of GBM. Genetic and molecular markers associated with PsP include: the MGMT promoter methylation [2, 5-12], Ki67 expression [5], IDH1 mutation [12], p53 mutation [13], interferon-regulatory factor 9 (IRF9) [14] and DNA repair protein(XRCC1) [15]. Nevertheless, the predictive value of these markers remains controversial [9, 16-18]. Hence, medical imaging techniques have been emerging as potential alternatives for PsP and TTP differentiation. In particular, magnetic resonance imaging (MRI) techniques that have been exploited for this task include diffusion-weighted imaging (DWI) [19-22], perfusion-weighted imaging (PWI) [19, 23], diffusion tensor imaging (DTI) [24-28], and three-dimensional echo planar spectroscopic imaging (3D-EPSI) [29]. However, these methods have achieved limited success for several reasons. Firstly, manual identification of lesion areas in medical images is subjective and costly. Next, the analysis of basic image features cannot adequately capture the subtle differences of PsP and TTP. Lastly, these methods have focused on using different image features to evaluate various MRI modalities, rather than developing an objective and automatic classification system for PsP and TTP.

With the development of deep learning, researchers have gradually discovered its potential for use in the field of image recognition [30-36]. Significant progress has been made using deep learning-based methods, especially deep convolutional neural network (CNN). This approach has greatly promoted the development of image classification and segmentation systems [37-39]. Compared with traditional pattern recognition methods, the greatest advantage of CNN for image classification is that it can learn image features automatically. This important advancement eliminates the complicated feature-engineering part of the traditional approach. Specifically, it is not necessary to study the local or global features while searching for what aspects best describe the characteristics of the image itself. Some approaches [40-44] based on CNN have achieved success in image classification tasks, but deep convolutional neural networks often confront a severe problem known as overfitting. Overfitting occurs because machine learning methods are required to train many learnable parameters, which in turn requires a large number of training samples. This issue becomes particularly tricky when the number of training samples is limited. In this domain, the problem is exemplified because medical images are difficult to obtain in large quantities due to both limited access and the inherent confidentiality of the subject matter. If there is an insufficient number of training samples, the deep model is often over-trained. Consequently, the model performs well during the training phase, but relatively poorly during the subsequent testing phase. Thus, it is essential to carry out a new and effective training strategy for deep learning models that solve the problem of overfitting.

A generative adversarial network (GAN) [45] is a deep learning architecture in which two neural networks compete against each other in a zero-sum game framework. A GAN can be regarded as a regularized learning scheme and can significantly alleviate the overfitting phenomenon. In particular, a GAN model consists of two parts: a generator and a discriminator. The generator produces synthetic images by imitating the original data distribution, whereas the discriminator is used to distinguish samples and classify them as real or generated [46-52]. In the learning stage of a GAN model [53], it is necessary to train the discriminator, D, to efficiently discern the source of input data as either genuine or fake. Simultaneously, the aim of the generator, G, is to create samples that are increasingly similar to the real images. Through the adversarial and competitive relationship between these two networks, when a limited number of training samples is used, the process of training the discriminator will not immediately succumb to overfitting. With the help of GANs, the problem of overfitting in deep learning can be significantly alleviated [51, 54, 55]. Although GANs have performed well in many fields, the adversarial nature of the method leads to problems of instability. In order to solve this, myriad techniques have been used to stabilize the training process. It has been discovered that a deep convolutional generative adversarial network plays an important role in eliminating instability problems.

The deep convolutional generative adversarial network (DCGAN), whose discriminator and generator are built on CNN, has achieved a high level of performance in image synthesis tasks [56]. Generator G, which takes a uniform noise distribution as input, can be reshaped into a multidimensional tensor. Discriminator D replaces pooling layers with stride convolutions on the basis of common CNN, and the activation functions are leaky rectified linear units. DCGAN shows potential in automatically learning data distribution characteristics, and it effectively alleviates the instability that comes with GAN training.

In this work, AlexNet [57] was chosen as the discriminator, and it was used to identify the features used for final classification. AlexNet is an architecture based on CNN that has proven success in scene classification tasks. It is recognized as an excellent basic level, automatic scene classification technology [58, 59]. While the typical pooling process is non-overlapping, AlexNet has, indeed, an overlapping pooling process. This contributes to a higher classification accuracy because more original information is retained. Suppose the kernel sizes are $z \times z$ and the stride is $s$ in each of the convolutional and deconvolutional layers. Upon setting $s=z$, traditional local pooling, as commonly employed in CNN, is obtained. If $s<z$ is set, overlapping pooling is obtained. Models with overlapping pooling are slightly more resistant to overfitting during training. In addition, feature fusion has been employed to combine high-layer features with low-layer features. This results in a more robust feature, which improves the final classification accuracy to some extent.

GANs represent a promising unsupervised learning scheme. However, so far, GANs have rarely been applied in the classification of PsP and TTP of GBM. We assert that for this kind of classification task, a GAN is an excellent choice because a GAN is an unsupervised learning architecture that depends on the generator to compensate for the shortage of training data. Thus, in this paper, we propose a model DC-AL GAN, which combines DCGAN with AlexNet to learn the representation of GBM images. The contributions of this work are the following.

1) The antagonism and competition between the discriminator and the generator lead to the extraction of highly refined features from the discriminator. The extraction of such features from PsP and TTP of GBM and the application of GANs to the classification of these features represent a novel application.

2) AlexNet was employed as the discriminator in this work, and the extracted features are used during the final classification. The typical pooling process is non-overlapping, while it is overlapping for AlexNet. This contributes to a higher classification accuracy because more original information is retained.

3) Feature fusion was implemented such that it combines features from both high and low layers, leading to features that are more precise. The result is an improvement in the final classification accuracy.

The framework of the proposed DC-AL GAN model is shown in Fig. 1

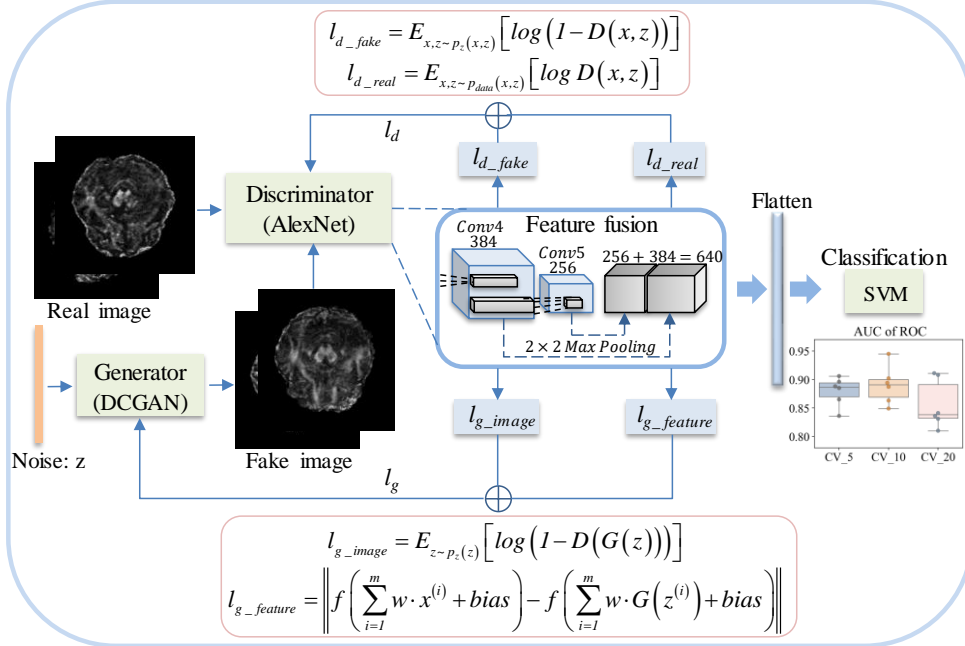

Fig. 1. Framework of the proposed DC-AL GAN model. The purpose of the generator is to produce samples that cannot be distinguished by the discriminator. The job of the discriminator is to classify data as either genuine or synthesized. Feature fusion shows the combination of high and low layer features. Classification expresses the boxplots of AUC of ROC when applying the extracted features to SVM with tenfold cross-validation.

## II. PRELIMINARIES

Section A introduces the data used in this paper. Section B describes the principles of the generative adversarial networks and summarizes the training procedure involving generator and discriminator. Section C shows the AlexNet architecture.

### A. Data Collection

A dataset composed of clinical records and longitudinal DTI data from 84 GBM patients (23 with PSP and 61 with TTP) was collected at the Wake Forest School of Medicine, Winston-Salem, NC. Each of these patients received a routine treatment, such as surgical resection followed by concurrent radiotherapy and chemotherapy with temozolomide. Each of the enrolled patients received a fixed dose of conformal radiotherapy (around 60 Gy). Along with radiotherapy and chemotherapy, each patients underwent DTI scanning (scanner: SIMCGEMR, GE Medical Systems) every two or three months for monitoring. The ultimate diagnosis results of PsP and TTP rely on the follow-up images and professional evaluation by experienced physicians.

### B. Generative Adversarial Networks

A generative adversarial network is a deep learning framework that is trained using an adversarial system. Synthetic data that adheres to the original distribution is generated to assist in the process of training. Unlike other deep learning models, a GAN consists of two parts: a generator and a discriminator. The generator synthesizes images by imitating the original data distribution, whereas the purpose of the discriminator is to distinguish a sample as either genuine or fake.

In GANs, the generator and discriminator perform training iteratively in separate, alternating rounds based on the minimax game-playing algorithm. First, the generator produces fake samples from random noise, which can initially fool the discriminator. Then the discriminator is supplied with both genuine and fake images, and learns to distinguish them. Throughout the process, the two parts update simultaneously, and this continues until the Nash equilibrium is satisfied.

The minimax game rules that both the generator and the discriminator obey the following optimization problem:

$$\min_G \max_D V(D,G) = E_{x \sim p_{data}(x)}[\log D(x)] \\ + E_{z \sim p_z(z)}[\log(1 - D(G(z)))] \quad (1)$$

D(x) represents the probability that x belongs to the class of real images, rather than fake samples. The primary goal of the discriminator is to calculate the probability at close to 1 when the input data is that of real images. When fake samples are considered, the purpose of the discriminator is to judge the data and learn what distinguishes them. Simultaneously, as D(G(z)) is close to 0, the generator aims to approach 1. This is essentially a minimax-style game between generator and discriminator.

In essence, during the learning phase for GANs, the aim is to train the model D such that it can effectively and efficiently discriminate the source of input as either real or fake. Simultaneously, model G aims to generate samples that are increasingly closer to real images.

### C. AlexNet Architecture

AlexNet is an architecture based on CNN that achieves convincing success in scene classification tasks and has proven to be an excellent basic level, automatic scene classification technology. AlexNet contains five convolutional layers and three fully-connected layers. Local response normalization is incorporated within AlexNet, to improve network generalization performance.

Assume that the input sample $X_i \in R^n$ denotes the input data, and $y_i \in \{1,..,K\}$ expresses the corresponding ground truth label for $X_i$. Further, suppose that the AlexNet model includes N layers, the weight combinations for the AlexNet architecture are $W = (W^{(1)},..,W^{(N)})$, and in this architecture, the relationships between the weight parameters and the filters are respectively shown in Eq. (2) and Eq. (3):

$$P^{(n)} = f(C^{(n)}) \quad (2)$$

$$C^{(n)} = W^{(n)} * P^{(n-1)} \quad (3)$$

In Eq. (2) and Eq. (3), $C^{(n)}$ refers to the convolved responses on the previous feature map; f() is the pooling function on C. P(W) refers to the output objective, which is defined in Eq. (4):

$$P(W) = \|w^{(out)}\|^2 + L(W, w^{(out)}) \quad (4)$$

where $\|w^{(out)}\|^2$ and $L(W, w^{(out)})$ are respectively the margin and squared hinge loss of the support vector machine (SVM) classifier. The overall loss of the output layer $L(W, w^{(out)})$ is shown in Eq. (5):

$$L(W, w^{(out)}) = \sum \left[1 - \langle w^{(out)}, f(P^{(N)}, y) - f(P^{(N)}, y_k) \rangle \right] \quad (5)$$

Eq. (5) represents the squared hinge loss of the prediction error. From the above description, it can be understood that, in Eq. (5), the AlexNet architecture learns the convolution kernels W. It can predict the label and give a strong push to having discriminative and sensible features at each layer. In this way, the overall goal of producing a good classification result in the output layer can be achieved. AlexNet architecture is shown in Fig. 2.

## III. DC-AL GAN ARCHITECTURE

In this section, the network architecture of the proposed model, DC-AL GAN, is described.

### A. Discriminator

AlexNet was utilized as the discriminator in this work. It contains five convolutional layers, and each of these follows a ReLU activation function. The extracted features mentioned in this work refer to output features from the discriminator's convolutional layer. Once training of the discriminator is complete, the output of the final layer in the discriminator model is regarded as the representation of the input image. The concept of feature fusion, which combines features from the different convolutional layers, has a positive effect on the final classification accuracy. The reason for the improvement is that more precise features are created during the fusion process. All of the convolution layers in the discriminator are subject to the ReLU activation function, and the slope is set to 0.2. During training, we utilized a stochastic gradient descent algorithm, where the batch size is set at 64. The Adam optimizer is used in the network, with the learning rate set to 0.0002, and momentum β1 as 0.5. The input images were scaled to [-1,1] before training the network. This is done in order to avoid any bias that is created by very large or very small numerical values. Another advantage is that it curbs numerical complexity during computation.

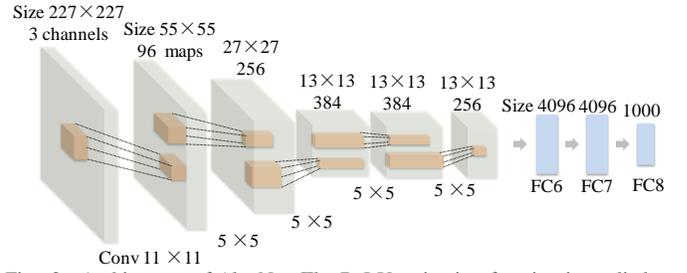

Fig. 2. Architecture of AlexNet. The ReLU activation function is applied to every convolutional layer.

When the discriminator is trained, the parameters of the generator are fixed. Optimizing the discriminator means maximization of the discriminate accuracy – therefore maximizing $V_D$ in Eq. (6):

$$V_D = E_{x,z \sim p_{data}(x,z)}\left[\log D(x,z)\right] + E_{x,z \sim p_z(x,z)}\left[\log(1 - D(x,z))\right] \quad (6)$$

When optimizing the discriminator, it is assumed that the generator has created fake samples. Optimizing the first item of Eq. (6) means that the output of the discriminator is maximized when inputting real images. This is because the prediction results of real images are expected to be close to 1. As for fake samples, optimization results should be minimized because the values are ideally close to 0. That is to say, the smaller the D(G(z)) is, the better the performance of the optimizer is. However, it is contradictory that when the first item is increasing, the second item decreasing. Therefore, we adjust the second item for 1-D(G(z)).

Given sufficient ability for the discriminator and generator to learn, GANs will have a global optimum. The optimum can be obtained by simple analysis. First, the discriminator can reach an optimum when the generator is fixed. Based on this value, an optimal function of the generator can be derived. It can be proven that the optimal function reaches a global minimum when generated distribution coincides with the actual data distribution.

When the generator is fixed, in order to achieve an optimal discriminator, the discriminator must be trained such that V(D, G) is maximized, as shown in the following Eq. (7):

$$\begin{aligned} V(D,G) &= \int_x p_{data}(x)\log(D(x))dx \\ &+ \int_z p_z(z)\log(1 - D(G(z)))dz \\ &= \int_x p_{data}(x)\log(D(x)) \\ &+ p_g(x)\log(1 - D(x))dx \end{aligned} \quad (7)$$

For any $(a,b) \in R^2 \setminus \{0,0\}$, function $y = a\log(y) + b\log(1 - y)$

achieves maximization at the point $\frac{a}{a+b}$, and the result follows after applying this function to Eq. (8):

$$D_G^*(x) = \frac{p_{data}(x)}{p_{data}(x) + p_g(x)} \quad (8)$$

That is, the optimal discriminator is as shown in Eq. (8) when the generator is fixed.

Jensen-Shannon (JS) divergence is a method based on Kullback-Leibler (KL) divergence that measures the similarity between two probability distributions. It has a symmetrical structure and a limited range of values. The definition for JS divergence is shown in Eq. (9):

$$JSD(P\|Q) = \frac{1}{2}D_{KL}(P\|M) + \frac{1}{2}D_{KL}(Q\|M) \quad (9)$$

In the above equality, $M = \frac{1}{2}(P+Q)$, P and Q represent probability distributions, respectively.

KL divergence is a method to measure how a probability distribution deviates from the standard probability distribution, as shown in Eq. (10):

$$D_{KL}(P\|Q) = \int_{-\infty}^{\infty} p(x) \log \frac{p(x)}{q(x)} dx \quad (10)$$

p and q represent the density of probability distributions P and Q, respectively.

When $p(x) = p_{data}(x)$, it is clear that $D_G^*(x) = \frac{1}{2}$. Applying Eq. (8) to Eq. (7), Eq. (11) follows where $m = \frac{p_{data}(x)}{p_{data}(x) + p(x)}$ and $n = \frac{p(x)}{p_{data}(x) + p(x)}$.

$$\begin{aligned}V(D_G^*, G) &= \int p_{data}(x)\log(m) + p(x)\log(n) dx \\ &= \int p_{data}(x)\log(m) + p(x)\log(n) dx \\ &\quad -\log 4 + \log 4 \int p(x) dx \\ &= \int p_{data}(x)\log(2m) + p(x)\log(2n) dx - \log 4 \quad (11)\\ &= D_{KL}\left(p_{data}(x) \left\| \frac{p_{data}(x) + p(x)}{2}\right.\right) \\ &\quad + D_{KL}\left(p(x) \left\| \frac{p_{data}(x) + p(x)}{2}\right.\right) - \log 4 \\ &= 2 \cdot JSD(p_{data}(x)\|p(x)) - \log 4\end{aligned}$$

The value of the JS divergence between two distributions is always non-negative. Also, JSD is zero when $p(x) = p_{data}(x)$. Therefore, the global minimum is $V(D_G^*, G) = -\log 4$. To clarify, when the generative distribution coincides with real data distribution, the function reaches a global minimum.

In the process of training the discriminator, $l_{d\_real}$ corresponds to the loss of real images. The output of the discriminator is expected to be as close to 1 as possible. This is because the result increases in accordance with the performance of the discriminator. The $l_{d\_real}$ is:

$$l_{d\_real} = E_{x,z \sim p_{data}(x,z)}\left[\log D(x,z)\right] \quad (12)$$

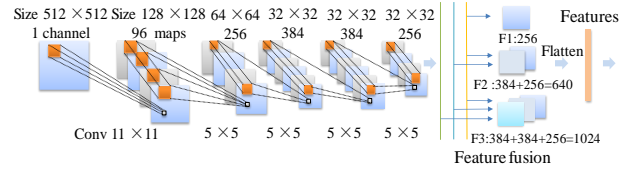

Fig. 3. AlexNet is used as the discriminator, which extracts more precise features by using feature fusion for the final classification. F1, F2 and F3 represent the features from the last one, the last two and the last three convolutional layer, respectively. The yellow, blue, and green lines represent the characteristics of the output of the final layer, the second last layer, and the third last layer, respectively.

The $l_{d\_fake}$ matches the loss of generated samples. The output should be as small as possible because an ideal discriminator will label fake samples as 0. The $l_{d\_fake}$ is:

$$l_{d\_fake} = E_{x,z \sim p_z(x,z)}\left[\log(1 - D(x,z))\right] \quad (13)$$

The discriminator loss ($l_d$) is comprised of $l_{d\_real}$ and $l_{d\_fake}$

$$l_d = l_{d\_real} + l_{d\_fake} \quad (14)$$

The architecture of the discriminator is shown in Fig. 3. The size of the input data is 512×512×1, where 512 represents the width and height of the input image, and 1 represents the gray image of 1 channel. The first layer of the AlexNet possesses 96 convolution kernels, where the kernel size is 11×11, and the stride is set to 4. The image size evolves into 128×128×96 through the convolutional layer. By performing 4×4 max pooling whose kernel size is 3×3 and stride is 2, we can produce feature maps with a size of 64×64×96. The input data of the second layer is of the size 64×64×96, the number of convolution kernels is 256, the kernel size is 5×5 and the stride is 1. Other parameter values are the same as those of the first layer. By performing convolution and max-pooling, we can produce 64×64×256 and 32×32×256 feature maps, respectively. The third and fourth layers only carried out convolution with 384 kernels, and the size of the data source remains unchanged after the ReLU activation function. The input of the fifth layer is 32×32×384, where other parameters are the same as the second layer. After convolution and max-pooling, the feature maps are 32×32×256 and 16×16×256, respectively. Finally, the features of different layers are fused and flattened for the output.

*B. Generator*

The generator is an improved network that is based on DCGAN. While the generator creates samples that are similar to the original data, the discriminator can learn more precise features from the input samples. This competitive process has the effect of each promoting the other. The input of the generator is a 100-dimensional uniform distribution z, which is then converted into a four-dimensional tensor. DC-AL GAN has three more convolutional layers than DCGAN. The seven convolutional layers are used to generate images of 512×512 pixels in size. The ReLU activation function is applied to all of the layers in the generator, in addition to the tanh function, which is used in the output layer. Batch normalization is employed in both generator and discriminator.

Regarding the generator and the mutual competition with the discriminator, it is expected that the output of the generated samples will approach one after being evaluated. The parameters of the discriminator are fixed when the generator is in the training phase. The purpose of training the generator is to boost the score of the synthetic samples, bringing them as close to 1 as possible. This means that D(G(z)) is increased, which affects $V_G$ in Eq. (15). In order to bring Eq. (15) into line with 1-D(G(z)), we adapt Eq. (15) to Eq. (16). In other words, optimizing the generator is to realize minimizing 1-D(G(z)), that is, minimizing $V_G$ in Eq. (16):

$$V_G = E_{z \sim p_z(z)}\left[\log\left(D(G(z))\right)\right] \quad (15)$$

$$V_G = E_{z \sim p_z(z)}\left[\log\left(1-D(G(z))\right)\right] \quad (16)$$

In the process of training GANs, the generator loss ($l_g$) is comprised of $l_{g\_image}$ and $l_{g\_feature}$. The $l_{g\_image}$ represents the deviation between the generated samples and the real images, as shown in Eq. (17). Reducing $l_{g\_image}$ by continuously training the network and adjusting parameters will enhance the generated images, bringing them closer to the real ones.

$$l_{g\_image} = E_{z \sim p_z(z)}\left[\log\left(1-D(G(z))\right)\right] \quad (17)$$

The $l_{g\_feature}$ describes the deviation between the output of the generator and that of the feature fusion layer. The greater the similarity between these, the more genuine the images appear.

Here, we set f(x) as the activation function in the discriminator. Both the convolutional layer and the pooling layer will possess a bias, and be activated such that the nonlinear characteristics can be better captured.

The formula of $l_{g\_feature}$ is shown in Eq. (18). Samples $\{x^1, x^2, \ldots, x^m\}$ come from the GBM dataset, whereas samples $\{z^1, z^2, \ldots, z^m\}$ originate from the random noise tensor, and w represents the convolution kernel.

$$l_{g\_feature} = \left\| f\left(\sum_{i=1}^m w \cdot x^{(i)} + bias\right) - f\left(\sum_{i=1}^m w \cdot G(z^{(i)}) + bias\right) \right\| \quad (18)$$

As aforementioned, the generator's loss contains two parts: $l_{g\_image}$ and $l_{g\_feature}$. The combination of these two kinds of loss increases the precision of the network during training. Consequently, the sample image produced by the generator is increasingly similar to the real image. In turn, the discriminator can extract more accurate features, and it improves the final classification accuracy. The functional expression of $l_g$ is shown in Eq. (19):

$$l_g = l_{g\_image} + l_{g\_feature} \quad (19)$$

The architecture used for the generator is shown in Fig. 4.

### C. Classification Using The SVM Algorithm

The Support Vector Machine (SVM) was first proposed by Cortes and Vapnik in 1995. It has many unique advantages in solving pattern recognition problems that are nonlinear, high-dimensional, and have a small sample size. The architecture can be extended to other machine learning cases such as function fitting. A linear SVM will create a situation where not all of the data will be partitioned; however, the majority of the data be correctly classified.

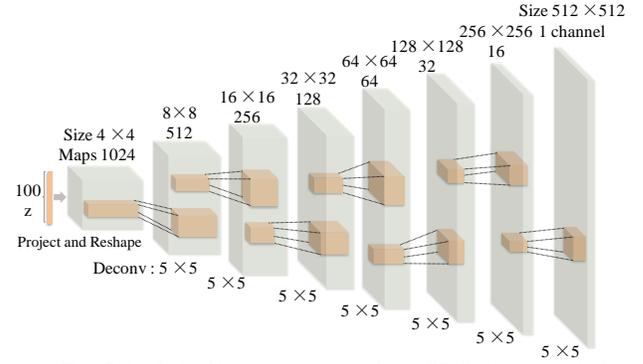

Fig. 4. The DC-AL GAN generator is used for GBM datasets, which is a modified network founded on deep convolutional generative adversarial networks. It contains seven convolutional layers, and can transform a 100 dimensional uniform tensor into 512×512 pixel images.

Tenfold cross-validation is used with the regular linear classifier L2-SVM for classification in DC-AL GAN. The features extracted from the discriminator are regarded as the input to the SVM for classification.

### IV. EXPERIMENTS

In this section, three sets of experiments are described to evaluate and compare the results of our method, DC-AL GAN, to other state-of-the-art methods. In the first experiment, the results of DC-AL GAN are compared with four related classification systems, including DCGAN, ResNet, DenseNet and VGG. The second experiment involves the analysis of classification results between sets of layers. Specifically, the last convolutional layer, the last two convolutional layers and the last three convolutional layers represented as F1, F2 and F3, respectively. In the third experiment, the best-performing combination from the second experiment (i.e., the features from the last two convolutional layers) is chosen and used for classification, and variations in k-fold cross-validation are compared with values k=5, 10 and 20, respectively.

In DC-AL GAN, tenfold cross-validation and regular linear classifier L2-SVM are used. This model uses TensorLayer, which is a library to facilitate deep learning (DL) and reinforcement learning (RL). It is an extension of Google TensorFlow. TensorLayer provides popular DL and RL modules that can be easily customized and assembled for tackling real-world machine learning problems.

### A. Comparison with Other Models

To analyze the classification performance of DC-AL GAN (DC-AL), we compared DCGAN, ResNet, DenseNet and VGG. All of these architectures were used in differentiating PsP and TTP of GBM. DC-AL GAN, DCGAN, ResNet, DenseNet and VGG reached an overall accuracy of 0.920, 0.844, 0.877, 0.873 and 0.862, respectively. Boxplots of the classification accuracy for each of the five models are shown in Fig. 5.

It can be seen that among the five methods, the median of DC-AL is much higher than the others. This means that classification using DC-AL is relatively stable. Because the average value is heavily influenced by extremes, it is sometimes unreasonable to use it as a measure. Preferably, the median is a better choice because it is less likely to be affected by very high or very low values.

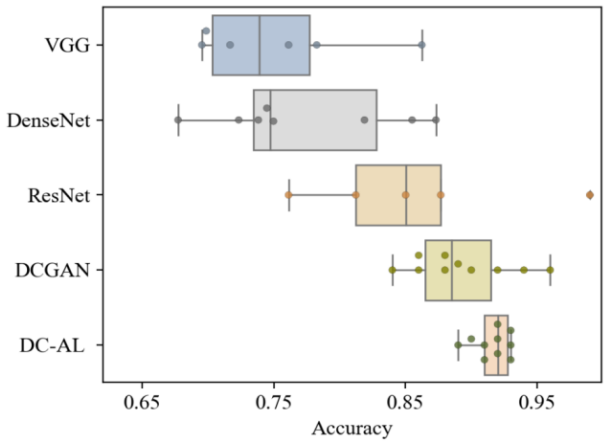

Fig. 5. Boxplots of the classification accuracy of the five models: DC-AL GAN (DC-AL), DCGAN, ResNet, DenseNet and VGG.

By examining the length of the box plots, it shows that the classification accuracy of DC-AL is relatively centralized and stable. In contrast, DenseNet is the worst performer. It can be concluded that the DC-AL GAN model proposed in this paper achieves higher classification accuracy when compared with the previous methods for the task at hand.

The sample images produced by the generator at different epochs are shown in Fig. 6. The last image in the figure consists of real samples, whereas the first seven images are the samples produced by the generator at different epochs. It is obvious that as the number of epochs increases, the generated samples gradually improve, becoming more like the genuine data. The result is a well-trained combination of G and D using only unlabeled samples. Also, D has learned the features from the data, which is beneficial for classification in the next steps.

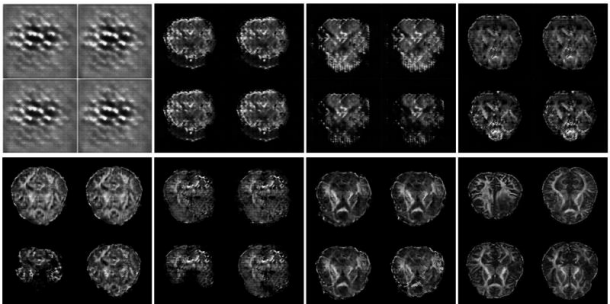

Fig. 6. DC-AL GAN-generated GBM images using unlabeled samples shown at different epochs. The first row represents epoch 0, epoch 5, epoch 10, and epoch 15 from left to right. The second row shows epoch 20, epoch 25, epoch 26 and the real image from the GBM dataset.

### B. Comparison of Classification Results Using Features Extracted from Different Convolutional Layer Sets

In this work, AlexNet has been used as the discriminator, which is also responsible for extracting the features used in the final classification. Besides, feature fusion has been employed to combine coarse, high-layer features with fine, low-layer features. The intention is to ensure that the network contains features with fine detail, yet retain high-level classification accuracy. Ultimately, the discriminator extracts features from the fusion layer. F1 represents the features from the final convolutional layer, F2 shows features from the last two layers, and F3 contains features from the last three layers.

The results of DC-AL GAN are illustrated in Fig. 7. The curves represent the accuracy for each of the feature sets. The overall accuracy achieved for each of F1, F2 and F3 is 0.893, 0.920 and 0.867, respectively. It is clear that the performance using set F2 is superior to that of the others. This means the combination of the last two convolutional layers results in the best combination. Post-analysis has shown that F3 performs poorly compared to F2. This is because the increase of the fusion layers leads to low-level features which are useless for classification.

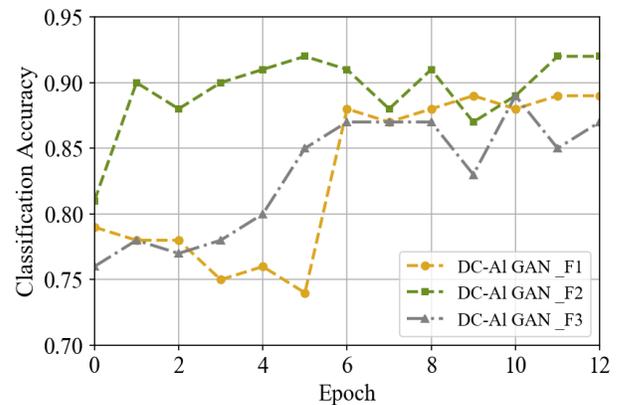

Fig. 7. The classification accuracy of DC-AL GAN with features F1, F2 and F3.

Table I contains additional statistics with respect the classification performance. The calculations are as follows: Sensitivity = TP/(TP + FN). Specificity = TN/(FP + TN). Precision = TP/(TP + FP). Where TP, FP, FN and TN refer to true positive, false positive, false negative and true negative, respectively.

TABLE I
COMPARISON OF CLASSIFICATION PERFORMANCE USING DIFFERENT FEATURES

| Methods | Sensitivity | Specificity | Precision |
|---|---|---|---|
| *DC-AL GAN_F1* | 0.912 | 0.747 | 0.881 |
| *DC-AL GAN_F2* | 0.976 | 0.883 | 0.945 |
| *DC-AL GAN_F3* | 0.929 | 0.833 | 0.920 |

Fig. 8 shows the confusion matrix for each of the three experiments. As seen in previous charts and tables, the model created using F2 is clearly the top performer.

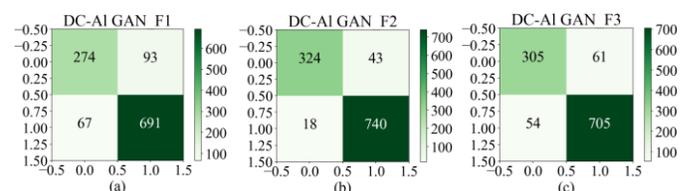

Fig. 8. (a)–(c) correspond to the confusion matrix of DC–AL GAN with features F1, F2 and F3, respectively.

## C. Comparison of Results for Differing Levels of Cross-Validation Using F2

Fig. 9(a) and Fig. 9(b) are boxplot representations of accuracy and AUC of ROC, respectively. Fig. 9(c), shows that the average accuracies were 0.902, 0.920 and 0.916 when CV=5, 10 and 20, respectively. These results indicate that this approach has a promising differentiation capability. The respective values for the AUC are 0.886, 0.947 and 0.931. In addition, as shown in Fig. 9(d), the p-values between performances with different repeated times are all greater than 0.05. Therefore, the difference in performance is insignificant and allows us to conclude that the classification system is stable.

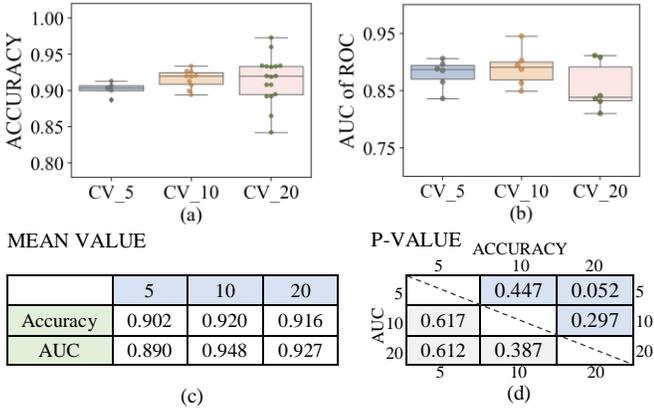

Fig. 9. (a) Accuracy corresponding to CV = 5, 10 and 20. (b) Area under the curve of receiver operating characteristics (ROC). (c) Mean values for accuracy and AUC. (d) P-values, calculated with t-test, for accuracy and AUC results.

In this section, the performance of different cross-validation values is compared through the receiver operating characteristic curves. The ROC curves correspond to 5, 10 and 20 CV repetitions, as shown in Fig. 10. The results show that this method achieves the best performance with 10 CV repetitions. In a reasonable range, appropriately increasing the number of folds of cross-validation can improve the generalization ability of the model and lead to better performance. However, the computational load should also be considered at the same time. For example, when CV is increased from 5 to 10, the performance of the model is obviously improved. When CV is increased from 10 to 20, the output results do not change significantly, but the calculation time is doubled. So in this paper, we adopt 10 CV repetitions. In Fig. 10(a)–(c), the areas under the ROC curves are close to 1, and the TPR values are greater than 0.8. The areas under the ROC curves of the 5 and 20 CV repetition are smaller than 10. Consequently, the experimental results confirm that this method has a better generalization capability.

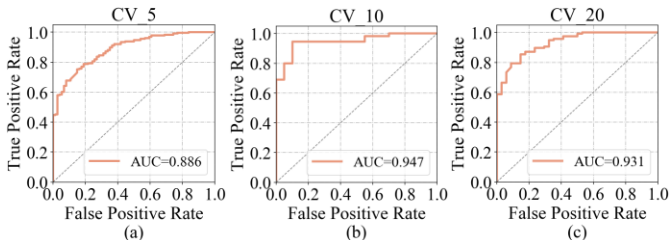

Fig. 10. ROC curves of these methods with 5, 10 and 20 CV repetitions.

## V. CONCLUSIONS AND FUTURE WORK

This paper introduces an unsupervised representation learning algorithm called DC-AL GAN. It is capable of learning interpretable representations, even from challenging GBM datasets. AlexNet is an integral component in the architecture where it is used as a discriminator to extract features. The results show that the discriminator can extract features that work effectively for classification. It does so by observing and analyzing the sample images created by the generator. Also, DC-AL GAN utilizes feature fusion by combining coarse, high-layer features with fine, low-layer features. This has shown to be beneficial in terms of classification performance. In summary, the experimental results have confirmed that DC-AL GAN achieves high accuracy on GBM datasets for PsP and TTP classification. Other possible future improvements to the work proposed in this paper include: optimizing the architecture of the generator to produce high-quality samples of images and classifying images in a semi-supervised manner to lower the demand for labeled data.